\documentclass[letterpaper, 10 pt, conference]{ieeeconf}  
\IEEEoverridecommandlockouts        
\makeatletter
\let\NAT@parse\undefined
\makeatother
\newcommand{\subparagraph}{}

\usepackage{algorithmic}
\usepackage{algorithm}
\usepackage[numbers,sort,comma]{natbib}
\usepackage{graphicx}
\usepackage{tabularx}
\usepackage{amsmath}
\usepackage{amssymb}
\usepackage{color}
\usepackage{graphicx,import}
\usepackage[compact]{titlesec}

\DeclareGraphicsExtensions{.pdf,.jpeg,.png}
\graphicspath{{figures/}}

\DeclareMathOperator*{\argmin}{arg\,min}

\titleformat{\subsubsection}[runin]{\normalsize\itshape}{\thesubsubsectiondis }{5pt}{}[:]

\titlespacing*{\subsubsection}{\parindent}{0ex plus 0.1ex minus 0.1ex}{5pt}

\title{\LARGE \bf Path Integral Guided Policy Search}

\author{Yevgen Chebotar$^{1}$ \and Mrinal Kalakrishnan$^{2}$ \and Ali Yahya$^{2}$ \and Adrian Li$^{2}$ \and Stefan Schaal$^{1}$ \and Sergey Levine$^{3}$ % 
\thanks{$^{1}$Yevgen Chebotar and Stefan Schaal are with the Department of Computer Science, University of Southern California, Los Angeles, CA 90089, USA. This research was conducted during Yevgen's internship at X, Mountain View, CA 94043, USA.
{\tt\small ychebota@usc.edu}}%
\thanks{$^{2}$Mrinal Kalakrishnan, Ali Yahya and Adrian Li are with X, Mountain View, CA 94043, USA.
{\tt\small kalakris@x.team}}%
\thanks{$^{3}$Sergey Levine is with Google Brain, Mountain View, CA 94043, USA.}
}

\begin{document}
\maketitle
\thispagestyle{empty}
\pagestyle{empty}

\begin{abstract}
We present a policy search method for learning complex feedback control policies that map from high-dimensional sensory inputs to motor torques, for manipulation tasks with discontinuous contact dynamics. We build on a prior technique called guided policy search (GPS), which iteratively optimizes a set of local policies for specific instances of a task, and uses these to train a complex, high-dimensional global policy that generalizes across task instances. We extend GPS in the following ways: (1) we propose the use of a model-free local optimizer based on path integral stochastic optimal control (PI$^2$), which enables us to learn local policies for tasks with highly discontinuous contact dynamics; and (2) we enable GPS to train on a new set of task instances in every iteration by using on-policy sampling: this increases the diversity of the instances that the policy is trained on, and is crucial for achieving good generalization. We show that these contributions enable us to learn deep neural network policies that can directly perform torque control from visual input.
We validate the method on a challenging door opening task and a pick-and-place task, and we demonstrate that our approach substantially outperforms the prior LQR-based local policy optimizer on these tasks. Furthermore, we show that on-policy sampling significantly increases the generalization ability of these policies.
\end{abstract}

\vspace{3pt}
\section{Introduction}
\vspace{1pt}

Reinforcement learning (RL) and policy search methods have shown considerable promise for enabling robots to automatically learn a wide range of complex skills \cite{TedrakeZS04,stoneICRA04,KoberMP08,DeisenrothRF11}, and recent results in deep reinforcement learning suggest that this capability can be extended to learn nonlinear policies that integrate complex sensory information and dynamically choose diverse and sophisticated control strategies \cite{lhphe-ccdrl-16,Levine:2016}.
However, applying direct deep reinforcement learning to real-world robotic tasks has proven challenging due to the high sample complexity of these methods. An alternative to direct deep reinforcement learning is to use guided policy search (GPS) methods, which use a set of local policies optimized on specific instances of a task (such as different positions of a target object) to train a global policy that generalizes across instances~\cite{Levine:2016}. In this setup, reinforcement learning is used only to train simple local policies, while the high-dimensional global policy, which might be represented by a deep neural network, is only trained with simple and scalable supervised learning methods.

The GPS framework can in principle use any learner to optimize the local policies. Prior implementations generally use a model-based method with local time-varying linear models and a local policy optimization based on linear-quadratic regulators (LQR)~\cite{Levine:2016}. We find that this procedure fails to optimize policies on tasks that involve complex contact switching discontinuities, such as door opening or picking and placing objects. In this paper, we present a method for local policy optimization using policy improvement with path integrals (PI$^2$)~\cite{TheodorouBS10}, and demonstrate the integration of this method into the GPS framework. We then enable GPS to train on new task instances in every iteration by extending the on-policy sampling approach proposed in recent work on mirror descent guided policy search (MDGPS) \cite{MontgomeryL16}. This extension enables robots to continually learn and improve policies on new task instances as they are experienced in the world, rather than training on a fixed set of instances in every iteration. This increases the diversity of experience and leads to substantially improved generalization, as demonstrated in our experimental evaluation.

\begin{figure}
  \centering
  \includegraphics[trim=0cm 0.0cm 0cm 0.0cm, clip=true,width=\columnwidth]{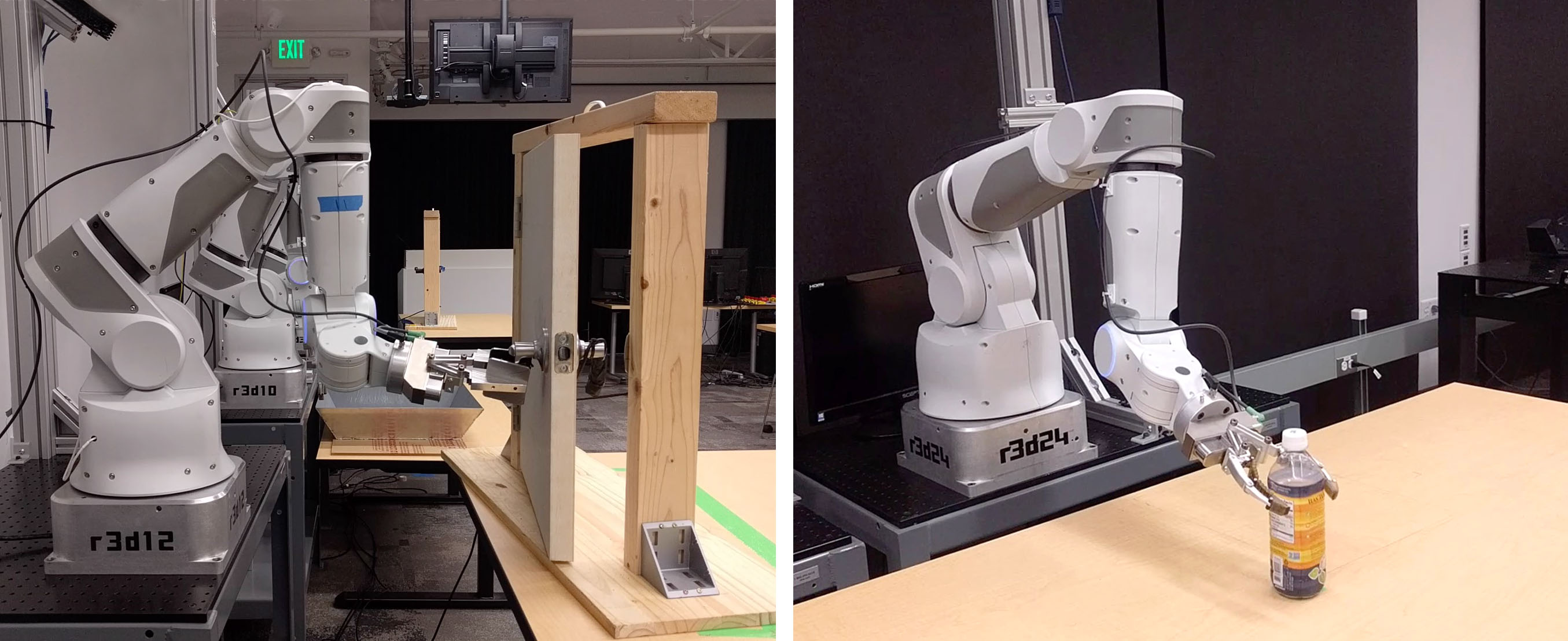}
  \vspace{-18pt}
  \caption{Door opening and pick-and-place using our path integral guided policy search algorithm. Door opening can handle variability in the door pose, while the pick-and-place policy can handle various initial object poses.}
  \label{fig:cover}
\end{figure}

We present real-world results for localizing and opening a door, as well as localizing and grasping an object and then placing it upright at a desired target location, shown in Figure~\ref{fig:cover}. Both tasks are initialized from demonstration and learned with our proposed path integral guided policy search algorithm, using deep visual features fed directly into the neural network policy. Our experimental results demonstrate that the use of stochastic local optimization with PI$^2$ enables our method to handle the complex contact structure of these tasks, and that the use of random instance sampling from the global policy enables superior generalization, when compared to prior guided policy search methods.

\vspace{3pt}
\section{Related Work}
\vspace{1pt}

Policy search methods have been used in robotics for a variety of tasks, such as manipulation \cite{PastorHAS09,DeisenrothRF11,ChebotarKP14}, playing table tennis \cite{KoberOP10} and ball-in-a-cup \cite{KoberMP08} games, regrasping \cite{regrasping}, and locomotion \cite{stoneICRA04,TedrakeZS04,EndoMMNC08}. Most of these works use carefully designed, specialized policies that either employ domain knowledge, or have a low number of parameters. It has been empirically observed that training high-dimensional policies, such as deep neural networks, becomes exceedingly difficult with standard model-free policy search methods \cite{policysearch}. Although deep reinforcement learning methods have made considerable progress in this regard in recent years \cite{slmja-trpo-15,lhphe-ccdrl-16}, their high sample complexity has limited their application to real-world robotic learning problems.

Guided policy search \cite{LevineK13} (GPS) seeks to address this challenge by decomposing policy search into trajectory optimization and supervised learning of a general high-dimensional policy. GPS was applied to various robotic tasks \cite{LevineA14,LevineWA15,Levine:2016}. However, the use of a model-based ``teacher'' to supervise the policy has placed considerable limitations on such methods. Most prior work has used LQR with fitted local time-varying linear models as the teacher \cite{LevineA14}, which can handle unknown dynamics, but struggles with problems that are inherently discontinuous, such as door opening: if the robot misses the door handle, it is difficult for a smooth LQR-based optimizer to understand how to improve the behavior. We extend GPS to tasks with highly discontinuous dynamics and non-differentiable costs by replacing the model-based LQR supervisor with PI$^2$, a model-free reinforcement learning algorithm based on stochastic optimal control \cite{TheodorouBS10}.

PI$^2$ has been successfully used to learn parameters of trajectory-centric policies such as dynamic movement primitives \cite{dmp} in grasping \cite{StulpTBS11}, pick-and-place \cite{StulpTS12} and variable impedance control tasks \cite{StulpBEMTS12}. 
% REVIEW: point 2
Compared to policy gradient methods, PI$^2$ does not compute a gradient, which is often sensitive to noise and large derivatives of the expected cost \cite{policysearch}. 
% END REVIEW: point  2
In \cite{KalakrishnanRPS11}, PI$^2$ is used to learn force/torque profiles for door opening and picking tasks to improve imperfect kinesthetic demonstrations. The policy is represented as an end-effector trajectory, and the motion is initialized from demonstration. 
In contrast, we use PI$^2$ to learn feedforward joint torque commands of time-varying linear-Gaussian controllers. In \cite{englert2016}, the robot also learns a policy for the door opening task initialized from demonstration. The training is divided into analytically learning a high-dimensional projection of the policy using provided models and Bayesian learning of a lower-dimensional projection for black-box objectives, such as binary success outcome of the task. In this work, we use the GPS framework to learn the high-dimensional policies using PI$^2$ as the teacher, which can also handle discontinuous cost functions. Furthermore, we use the controls learned by PI$^2$ on several task instances (e.g. several door positions) to supervise the training of a single deep neural network policy that can succeed for various door poses using visual input from the robot's camera.

Stochastic policy search methods can be improved by limiting the information loss between updates, by means of a KL-divergence constraint \cite{PetersMA10}. In this work, we similarly constrain the KL-divergence between PI$^2$ updates, in a framework similar to \cite{lioutikov} and \cite{GomezKPN14}. In \cite{Hoof0N15}, the authors propose to learn high-dimensional sensor-based policies through supervised learning using the relative entropy method to reweight state-action samples of the policy. While the goal of learning high-dimensional nonlinear policies is similar to our work, we optimize the individual instance trajectories separately, and then combine them into a single policy with supervised learning. As shown in our simulated experimental evaluation, this substantially improves the effectiveness of the method and allows us to tackle more complex tasks.
We also extend guided policy search by choosing new random instances at each iteration, based on the on-policy sampling technique proposed in \cite{MontgomeryL16}, which substantially improves the generalization of the resulting policy.

Our deep neural network policies directly use visual features from the robot's camera to perform the task. The features are learned automatically on a pose detection proxy task, using an improved version of the spatial feature points architecture \cite{Levine:2016} based on convolutional neural networks (CNNs) \cite{Fuk80,Schmidhuber15}. 
% REVIEW: point 3
%In contrast to visual servoing \cite{ECR92, WilsonHB96}, our approach does not require any manually engineered feature points, feedback controllers or camera calibration, as our visuomotor policies are entirely learned from data. 
% ENDREVIEW: point 3
In \cite{Koutnik14}, visual and control layers of a racing video game policy are learned separately using neuroevolution. Using pre-trained visual features enables efficient learning of smaller controllers with RL. In our work, visual features are pre-trained based on object and robot end-effector poses. By combining visual pre-training, initialization from kinesthetic demonstrations, and global policy sampling with PI$^2$, we are able to learn complex visuomotor behaviors for contact-rich tasks with discontinuous dynamics, such as door opening and pick-and-place.

\vspace{3pt}
\section{Background}
\vspace{1pt}

\label{sec:background}
In this section, we provide background on guided policy search (GPS) \cite{LevineA14} and describe the general framework of model-free stochastic trajectory optimization using policy improvement with path integrals (PI$^2$) \cite{TheodorouBS10}, which serve as the foundations for our algorithm.

\subsection{Guided Policy Search}
\label{sec:gps}
The goal of policy search methods is to optimize parameters $\theta$ of a policy $\pi_\theta(\mathbf{u}_t | \mathbf{x}_t)$, which defines a probability distribution over robot actions $\mathbf{u}_t$ conditioned on the system state $\mathbf{x}_t$ at each time step $t$ of a task execution. Let $\tau = (\mathbf{x}_1, \mathbf{u}_1, \dots,  \mathbf{x}_T, \mathbf{u}_T)$ be a trajectory of states and actions. Given a task cost function $l(\mathbf{x}_t, \mathbf{u}_t)$, we define the trajectory cost $l(\tau)=\sum_{t=1}^T l(\mathbf{x}_t, \mathbf{u}_t)$. The policy optimization is performed with respect to the expected cost of the policy:
\vspace{-5pt}
\[
J(\theta) = E_{\pi_\theta}\left[l(\tau)\right] = \int l(\tau) p_{\pi_\theta} (\tau) d\tau,
\vspace{-4pt}
\]
where $p_{\pi_\theta} (\tau)$ is the policy trajectory distribution given the system dynamics $p\left(\mathbf{x}_{t+1} | \mathbf{x}_{t}, \mathbf{u}_{t}\right)$:
\vspace{-3pt}
\[
p_{\pi_\theta} (\tau) = p(\mathbf{x}_1) \prod_{t=1}^{T} p\left(\mathbf{x}_{t+1} | \mathbf{x}_{t}, \mathbf{u}_{t}\right) \pi_\theta(\mathbf{u}_t | \mathbf{x}_t).
\vspace{-4pt}
\]
Standard policy gradient methods optimize $J(\theta)$ directly with respect to the parameters $\theta$ by estimating the gradient $\nabla_\theta J(\theta)$. The main disadvantage of this approach is that it requires large amounts of policy samples and it is prone to fall into poor local optima when learning high-dimensional policies with a large number of parameters \cite{policysearch}. 

Guided policy search (GPS) introduces a two-step approach for learning high-dimensional policies by leveraging advantages of trajectory optimization and supervised learning. Instead of directly learning the policy parameters with reinforcement learning, a trajectory-centric algorithm is first used to learn simple controllers $p(\mathbf{u}_t | \mathbf{x}_t)$ for trajectories with various initial conditions of the task. We refer to these controllers as \textit{local policies}. 
In this work, we employ time-varying linear-Gaussian controllers of the form $p(\mathbf{u}_t | \mathbf{x}_t) = \mathcal{N}(\mathbf{K}_{t} \mathbf{x}_t + \mathbf{k}_{t}, \mathbf{C}_{t})$ to represent these local policies.

After optimizing the local policies, the optimized controls from these policies are used to create a training set for learning a complex high-dimensional \textit{global policy} in a supervised manner. Hence, the final global policy generalizes to the initial conditions of multiple local policies and can contain thousands of parameters, which can be efficiently learned with supervised learning. Furthermore, while trajectory optimization might require the full state $\mathbf{x}_t$ of the system to be known, it is possible to only use the observations $\mathbf{o}_t$ of the full state for training a global policy $\pi_\theta(\mathbf{u}_t | \mathbf{o}_t)$. In this way, the global policy can predict actions from raw observations at test time \cite{Levine:2016}.

Supervised training does not guarantee that the global and local policies match as they might have different state distributions. In this work, we build on MDGPS, which uses a constrained formulation of the local policy objective \cite{MontgomeryL16}:
\vspace{-2pt}
\[
\min_{\theta, p} E_p[l(\tau)] \,\, \mbox{s.t.}\,\, D_\text{KL} \left(p(\tau) \|\, \pi_{\theta} (\tau) \right) \leq \epsilon, 
\vspace{-3pt}
\]
where $D_\text{KL} \left(p(\tau) \|\, \pi_{\theta} (\tau) \right)$ can be computed as: 
\vspace{-2pt}
\[
D_\text{KL} \left(p(\tau) \|\, \pi_{\theta} (\tau) \right) = \sum_{t=1}^T E_{p} \left [ D_\text{KL} \left(p(\mathbf{u}_t | \mathbf{x}_{t}) \|\,  \pi_\theta (\mathbf{u}_t | \mathbf{x}_{t})  \right) \right].
\vspace{-2pt}
\]
The MDGPS algorithm alternates between solving the constrained optimization with respect to the local policies $p$, and minimizing the KL-divergence with respect to the parameters of the global policy $\theta$ by training the global policy on samples from the local policies with supervised learning.

In prior GPS work, the local policies were learned by iteratively fitting time-varying linear dynamics to data, and then improving the local policies with a KL-constrained variant of LQR. While this allows for rapid learning of complex trajectories, the smooth LQR method performs poorly in the presence of severe discontinuities. In this work, we instead optimize the local policies using the model-free PI$^2$ algorithm, which is described in the next section.

\subsection{Policy Improvement with Path Integrals}
\label{sec:pi2}
PI$^2$ is a model-free RL algorithm based on stochastic optimal control 
and statistical estimation theory. Its detailed derivation can be found in \cite{TheodorouBS10}. We outline the method below and describe its application to learning feedforward commands of time-varying linear-Gaussian controllers.

The time-varying linear-Gaussian controllers we use to represent the local policies are given by $p(\mathbf{u}_t | \mathbf{x}_t) = \mathcal{N}(\mathbf{K}_t \mathbf{x}_t  + \mathbf{k}_t, \mathbf{C}_t)$. They are parameterized by the feedback gain matrix $\mathbf{K}_t$, the feedforward controls $\mathbf{k}_t$, and the covariance $\mathbf{C}_t$.
In this work, we employ PI$^2$ to learn only the feedforward terms and covariances. After initialization, e.g. from human demonstrations, the feedback part is kept fixed throughout the optimization. In this manner, we can keep the number of learned parameters relatively small and be able to learn the local policies with a low number of policy samples.

Each iteration of PI$^2$ involves generating $N$ samples by running the current policy on the robot. After that, we  compute the cost-to-go $S_{i,t}$ and probabilities $P_{i,t}$ for each sample $i \in 1 \dots N$ and for each time step $t$:
\vspace{-5pt}
\[
S_{i,t} = S(\tau_{i,t}) = \sum^T_{j=t} l(\mathbf{x}_{i,j}, \mathbf{u}_{i,j}), \,\,\, P_{i,t} = \frac{ e^{-\frac{1}{\eta} S_{i,t}}}{\sum_{i=1}^N e^{-\frac{1}{\eta} S_{i,t}}},
\vspace{-3pt}
\]
where $l(\mathbf{x}_{i,j}, \mathbf{u}_{i,j})$ is the cost of sample $i$ at time $j$. 
The probabilities follow from the Feynman-Kac theorem applied to stochastic optimal control \cite{TheodorouBS10}.
The intuition is that the trajectories with lower costs receive higher probabilities, and the policy distribution shifts towards a lower cost trajectory region. The costs are scaled by $\eta$, which can be interpreted as the temperature of a soft-max distribution. 

After computing the new probabilities $P_{i,t}$, the policy is updated according to a weighted maximum-likelihood estimate of the policy parameters, which are the means and covariances of the sampled feedforward controls $\mathbf{\bar{k}}_{i,t}$:
%REVIEW remove 'new' and add bar above the sampled controls to save some space
%\vspace{-4pt}
%\[
%\mathbf{k}_t^{new} = \sum_{i=1}^N P_{i,t} \mathbf{k}_{i,t} 
%\]
%\vspace{-13pt}
%\[
%\mathbf{C}_t^{new} = \sum_{i=1}^N P_{i,t} (\mathbf{k}_{i,t} - %\mathbf{k}_t^{new})(\mathbf{k}_{i,t} - \mathbf{k}_t^{new})^\top
%\vspace{-4pt}
%\]
\vspace{-5pt}
\[
\mathbf{k}_t = \sum_{i=1}^N P_{i,t} \mathbf{\bar{k}}_{i,t}, ~\,
\mathbf{C}_t = \sum_{i=1}^N P_{i,t} (\mathbf{\bar{k}}_{i,t} - \mathbf{k}_t)(\mathbf{\bar{k}}_{i,t} - \mathbf{k}_t)^\top
\vspace{-3pt}
\]
Here, we adapt the approach described in \cite{StulpS12} and update the covariance matrices based on the sample probabilities. This has the advantage of automatically determining the exploration magnitude for each time step, instead of setting it to a constant as in the original PI$^2$ derivation.

\vspace{3pt}
\section{Path Integral Guided Policy Search}

In this section, we first describe how to use PI$^2$ as the local policy optimizer in guided policy search, and then
introduce global policy sampling as a way to train policies on randomized task instances using raw observations. Finally, we describe the neural network architecture used to represent our visuomotor global policies and the pre-training procedure that we use to learn visual features.

\subsection{PI$^2$ for Guided Policy Search}
\label{pigps}
Using the time-varying linear-Gaussian representation of the local policies makes it straightforward to incorporate PI$^2$ into the GPS framework from Section~\ref{sec:gps}. 

In previous work \cite{LevineA14}, the optimization of local policies with LQR was constrained by the KL-divergence between the old and updated trajectory distributions to ensure steady policy improvement. Similar types of constraints were proposed in prior policy search work \cite{PetersMA10}. When optimizing with PI$^2$, we similarly want to limit the change of the policy and hence, avoid sampling too far from unexplored policy regions to gradually converge towards the optimal policy. The policy update step can be controlled by varying the temperature $\eta$ of the soft-max probabilities $P_{i,t}$. If the temperature is too low, the policy might converge quickly to a suboptimal solution. If $\eta$ is too high, the policy will converge slowly and the learning can require a large number of iterations. Prior work set this parameter empirically \cite{TheodorouBS10}.

In this work, we employ an approach similar to relative entropy policy search (REPS) \cite{PetersMA10} to determine $\eta$ based on the bounded information loss between the old and updated policies. For that, the  optimization goal of the local policy $p\left (\mathbf{u}_t | \mathbf{x}_t \right)$ is augmented with a KL-divergence constraint against the old policy $\bar{p}\left (\mathbf{u}_t | \mathbf{x}_t \right)$:
\vspace{-2pt}
  \[
 \min_{p} E_p[l(\tau)] \,\,\, \mbox{s.t.} \,\,\, D_\text{KL} \left(p\left (  \mathbf{u}_t  | \mathbf{x}_t \right) \|\, \bar{p}\left (  \mathbf{u}_t  | \mathbf{x}_t \right) \right) \leq \epsilon,
 \vspace{-3pt}
\]
where $\epsilon$ is the maximum KL-divergence between the old and new policies. The Lagrangian of this problem depends on $\eta$:
\[
\mathcal{L}_p(p, \eta) = E_p[l(\tau)] + \eta \left [D_\text{KL} \left(p\left (\mathbf{u}_t  | \mathbf{x}_t \right) \|\, \bar{p}\left(\mathbf{u}_t| \mathbf{x}_t \right) \right) - \epsilon \right]
\]
We compute the temperatures $\eta_t$  separately for each time step by optimizing the dual function $g(\eta_t)$ with respect to the cost-to-go of the policy samples:
\vspace{-4pt}
\[
g(\eta_t) = \eta_t \epsilon + \eta_t \log \frac{1}{N} \sum_{i=1}^N \left [ e^{-\frac{1}{\eta_t} S_{i,t}} \right ].
\vspace{-3pt}
\]
The derivation of the dual function follows \cite{PetersMA10}, but performed at each time step independently as in \cite{lioutikov}.  

By replacing $\bar{p}\left(\mathbf{u}_t| \mathbf{x}_t \right)$ with the current global policy $\pi_{\theta} (\mathbf{u}_t | \mathbf{x}_t)$, we obtain the mirror gradient descent formulation of GPS (MDGPS) \cite{MontgomeryL16}:
\vspace{-2pt}
\[
 \min_{p} E_p[l(\tau)] \,\,\, \mbox{s.t.} \,\,\, D_\text{KL} \left(p\left (  \mathbf{u}_t  | \mathbf{x}_t \right) \|\, \pi_\theta\left (  \mathbf{u}_t  | \mathbf{x}_t \right) \right) \leq \epsilon.
 \vspace{-4pt}
\]
We can therefore use the same relative entropy optimization procedure for $\eta$ to limit deviation of the updated local policy from the old global policy, which is also used in our global policy sampling scheme as described in the next section.

\subsection{Global Policy Sampling}

In the standard GPS framework, the robot trajectories are always sampled from the local policies. This has a limitation of being constrained to a fixed set of task instances and initial conditions for the entire learning process. However, if we use PI$^2$ with a constraint against the previous global policy, we in fact do not require a previous local policy for the same instance, and can therefore sample new instances at each iteration. This helps us to train global policies with a better generalization to various task conditions.

\begin{algorithm}[t]
\begin{algorithmic}[1]
\FOR{iteration $k \in \{1,\dots,K\}$}
    \STATE{ Generate samples $\mathcal{D} = \{\tau_{i}\}$ by running noisy $\pi_{\theta}$ on each randomly sampled instance}
    \STATE{Perform one step of optimization with PI$^2$ independently on each instance:\\ 
    $\min_{p} E_p[l(\tau)] \,\, \mbox{s.t.}\,\, D_{KL} \left(p(\mathbf{u}_t | \mathbf{x}_t) \|\, \pi_{\theta} (\mathbf{u}_t | \mathbf{x}_t) \right) \leq \epsilon$}
    \STATE{Train global policy with optimized controls using supervised learning:\\
    $ \pi_\theta \gets \argmin_\theta \sum_{i,t} D_\text{KL} \left(\pi_\theta (\mathbf{u}_t | \mathbf{x}_{i,t}) \|\, p(\mathbf{u}_t | \mathbf{x}_{i,t}) \right)$}
\ENDFOR
\end{algorithmic}
\caption{MDGPS with PI$^2$ and Global Policy Sampling}
\label{algo:pigps}
\end{algorithm}

The approach, summarized in Algorithm \ref{algo:pigps}, is similar to MDGPS and inherits its theoretical underpinnings. However, unlike standard MDGPS, we sample new instances (e.g. new poses of the door) at each iteration. We then perform a number of rollouts for each instance using the global policy $\pi_\theta(\mathbf{u}_t|\mathbf{o}_t)$, with added noise.
These samples are used to perform a single optimization step with PI$^2$, independently for each instance. As we use samples from the global policy, the optimization is now constrained against KL-divergence to the old global policy and can be solved through optimization of the temperature $\eta$, as described in the previous section.
After performing one step of local policy optimization with PI$^2$, the updated controls are fed back into the global policy with supervised learning. The covariance of the global policy noise is updated by averaging the covariances of the local policies at each iteration:
%REVIEW: the covariances are averaged across samples and time, not only across samples as in the previous version
$\mathbf{C}_{\pi_\theta} = \sum_{i, t} \mathbf{C}_{p_i,t} / (NT)$. 
%END REVIEW

Our approach is different from direct policy gradient on the parameters of the global policy. Although we sample from a high-dimensional global policy, the optimization is still performed in a lower-dimensional action space with a trajectory-centric algorithm that can take advantage of the local trajectory structure. Consequently, the global policy is guided by this optimization with supervised learning.

\subsection{Global Policy Initialization}

It is important to note that, especially for real robot tasks, it is often not safe or efficient to start with an uninitialized or randomly initialized global policy, particularly when using very general representations like neural networks. Therefore, in our work we initialize the global policies by performing several iterations of standard GPS with local policy sampling using PI$^2$ on a fixed set of task instances. In this case, the cost-to-go $S_{i,t}$ in PI$^2$ is augmented with a KL-divergence penalty against the global policy as described in \cite{Levine:2016} (Appendix A), but the optimization is performed using the KL-divergence constraint against the previous local policy. We also initialize the local policies with kinesthetic teaching, to provide the algorithm with the overall structure of the task at the start of training. After initialization, the global policy can already occasionally perform the task, but is often overfitted to the initial task instances. By training further with global policy sampling on random task instances, we can greatly increase the generalization capability of the policy, as demonstrated in our experimental evaluation.

\subsection{Learning Visuomotor Policies}
\label{sec:visuomotor}

\begin{figure*}[t!]
\centering
\includegraphics[trim=0pt 97pt 2pt 111pt, clip=true,width=0.9\textwidth]{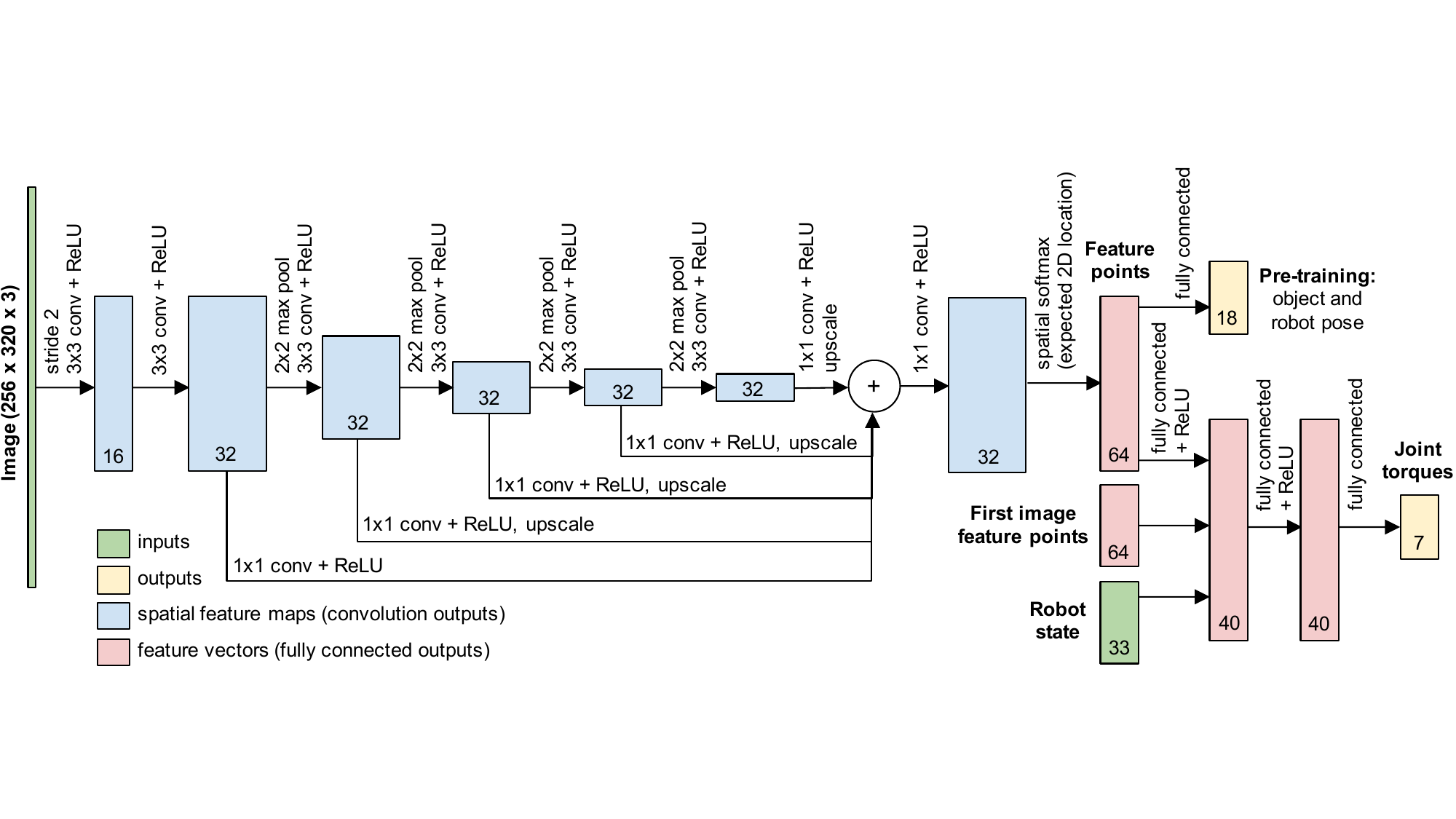}
\vspace{-9pt}
\caption{The architecture of our neural network policy. The input RGB image is passed through a 3x3 convolution with stride 2 to generate 16 features at a lower resolution. The next 5 layers are 3x3 convolutions followed by 2x2 max-pooling, each of which output 32 features at successively reduced resolutions and increased receptive field. The outputs of these 5 layers are recombined by passing each of them into a 1x1 convolution, converting them to a size of 125x157 by using nearest-neighbor upscaling, and summation (similar to~\cite{tompson2014joint}). A final 1x1 convolution is used to generate 32 feature maps. The spatial soft-argmax operator~\cite{Levine:2016} computes the expected 2D image coordinates of each feature. A fully connected layer is used to compute the object and robot pose from these expected 2D feature coordinates for pre-training the vision layers. The feature points for the current image are concatenated with feature points from the image at the first timestep as well as the 33-dimensional robot state vector, before being passed through two fully connected layers to produce the output joint torques.}
\label{fig:nn}
\vspace{-12pt}
\end{figure*}

Our goal is to use the proposed path integral guided policy search algorithm to learn policies for complex object manipulation skills. Besides learning how to perform the physical motions, these policies must also interpret visual inputs to localize the objects of interest. To that end, we use learned deep visual features as observations for the policy. The visual features are produced by a convolutional neural network, and the entire policy architecture, including the visual features and the fully-connected motor control layers, is shown in Figure~\ref{fig:nn}. Our architecture resembles prior work~\cite{Levine:2016}, with the visual features represented by feature points produced via a spatial softmax applied to the last convolutional response maps. Unlike prior work, our convolutional network includes pooling and skip connections, which allows the visual features to incorporate information at various scales: low-level, high-resolution, local features as well as higher-level features with larger spatial context. This serves to limit the amount of computation performed at high resolutions while still generating high-resolution features, enabling evaluation of this deep model at camera frame rates.
\\
We train the network in two stages. First, the convolutional layers are pre-trained with a proxy pose detection objective. To create data for this pre-training phase, we collect camera images while manually moving the object of interest (the door or the bottle for the pick-and-place task) into various poses, and automatically label each image by using a geometry-based pose estimator based on the point pair feature (PPF) algorithm~\cite{stefanppf}. We also collect images of the robot learning the task with PI$^2$ (without vision), and label these images with the pose of the robot end-effector obtained from forward kinematics. Each pose is represented as a 9-DoF vector, containing the positions of three points rigidly attached to the object (or robot), represented in the world frame. The convolutional layers are trained using stochastic gradient descent (SGD) with momentum to predict the end-effector and object poses, using a standard Euclidean loss. The fully connected layers of the network are then trained using path integral guided policy search to produce the joint torques, while the weights in the convolutional layers are frozen. Since our model does not use memory, we include the first camera image feature points in addition to the current image to allow the policy to remember the location of the object in case of occlusions by the arm. In future work, it would be straightforward to also fine-tune the convolutional layers end-to-end with guided policy search as in prior work~\cite{Levine:2016}, but we found that we could obtain satisfactory performance on the door and pick-and-place tasks without end-to-end training of the vision layers.

\section{Experiments}

The first goal of our experiments is to compare the performance of guided policy search with PI$^2$ to the previous LQR-based variant on real robotic manipulation tasks with discontinuous dynamics and non-differentiable costs. We evaluate the algorithms on door opening and pick-and-place tasks. The second goal is to evaluate the benefits of global policy sampling, where new random instances (e.g. new door poses) are chosen at each iteration. To that end, we compare the generalization capabilities of policies trained with and without resampling of new instances at each iteration. Finally, we present simulated comparisons to evaluate the design choices in our method, including the particular variant of PI$^2$ proposed in this work.

\subsection{Experimental setup}

\begin{figure*}[t]
\centering
\includegraphics[trim=0 0 0 0, clip=true,width=\textwidth]{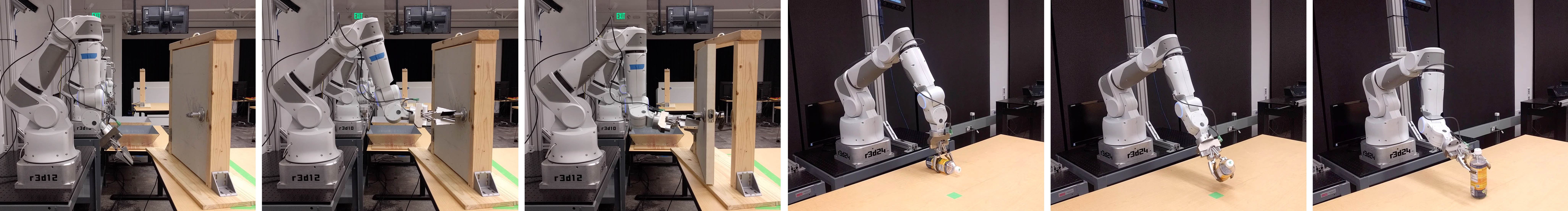}
\vspace{-19pt}
\caption{Task setup and execution. \textit{Left:} door opening task. \textit{Right:} pick-and-place task. For both tasks, the pose of the object of interest (door or bottle) is randomized, and the robot must perform the task using monocular camera images from the camera mounted over the robot's shoulder.}
\label{fig:tasks}
\vspace{-14pt}
\end{figure*}

We use a lightweight 7-DoF robotic arm with a two finger gripper, and a camera mounted behind the arm looking over the shoulder (see Figure~\ref{fig:tasks}). The input to the policy consists of monocular RGB images, with depth data used only for ground truth pose detection during pre-training. The robot is controlled at a frequency of 20Hz by directly sending torque commands to all seven joints. We use different fingers for the door opening and pick-and-place tasks.

\subsubsection{Door opening} In the door opening task, the goal of the robot is to open the door as depicted in Figure~\ref{fig:tasks} on the left.
The cost is based on an IMU sensor attached to the door handle. The desired IMU readings, which correspond to a successfully opened door, are recorded during kinesthetic teaching of the opening motion from human demonstration. We additionally add joint velocity and control costs multiplied by a small constant to encourage smooth motions.

\subsubsection{Pick-and-place}
The goal of the pick-and-place task is to pick up a bottle and place it upright at a goal position, as shown in Figure~\ref{fig:tasks} on the right.
The cost is based on the deviation of the final pose of the bottle from the goal pose. We use object detection with PPF \cite{stefanppf} to determine the final pose to evaluate the cost. This pose is only used for cost evaluation, and is not provided to the policy. The goal pose is recorded during the human demonstration of the task. The cost is based on the quadratic distance between three points projected into the goal and final object pose transformations, and we again add small joint velocity and control costs.

\subsection{Single-instance comparisons}

\subsubsection{Door opening}
\begin{figure}
  \centering
  \includegraphics[trim=0cm 0pt 0cm 0pt, clip=true,width=0.987\columnwidth]{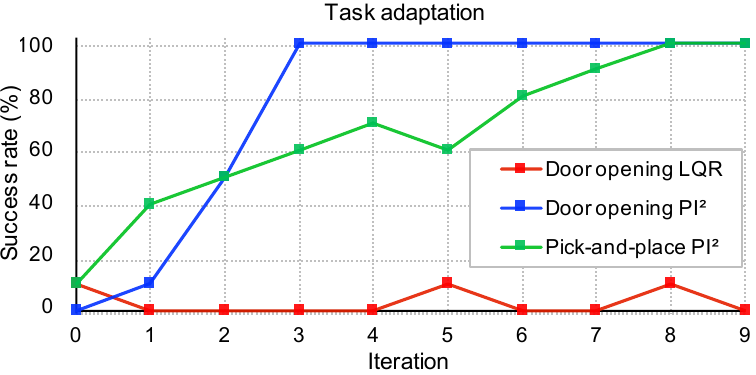}
  \vspace{-8pt}
  \caption{Task adaptation success rates over the course of training with PI$^2$ and LQR for single instances of door opening and pick-and-place tasks. Each iteration consists of 10 trajectory samples.}
  \label{fig:task_adapt}
\end{figure}

We first evaluate how our method performs on a single task instance (a single position of the door) without vision, so as to construct a controlled comparison against prior work. After recording the door opening motion from demonstration, the door is displaced by 5cm away from the robot. The robot must adapt the demonstrated trajectory to the new door position and recover the opening motion. The linear-Gaussian policy has 100 time steps with torque commands for each of the joints. We add control noise at each of the time steps to start exploration around the initial trajectory. The amount of the initial noise is set such that the robot can touch the door handle at least 10\% of the time, since the only feedback about the success of the task comes from the IMU readings on the door handle.

We compare the PI$^2$ variant of GPS to the more standard LQR version \cite{Levine:2016} over 10 iterations, with 10 sampled trajectories at each iteration. Figure~\ref{fig:task_adapt} shows the success rates for opening the displaced door using LQR and PI$^2$. In the initial trials, the robot is able to either open the door once or touch the door handle a few times. After the first two policy updates, PI$^2$ could open the door in 50\% of the samples, and after three updates the door could be opened consistently. LQR could not handle the non-linearity of the dynamics due to the contacts with the door and discontinuity in the cost function between moving and not moving the door handle.
This shows that the model-free PI$^2$ method is better suited for tasks with intermittent contacts and non-differentiable costs.

\subsubsection{Pick-and-place}
\begin{figure}
  \centering
  \includegraphics[trim=0cm 0 0cm 0, clip=true,width=\columnwidth]{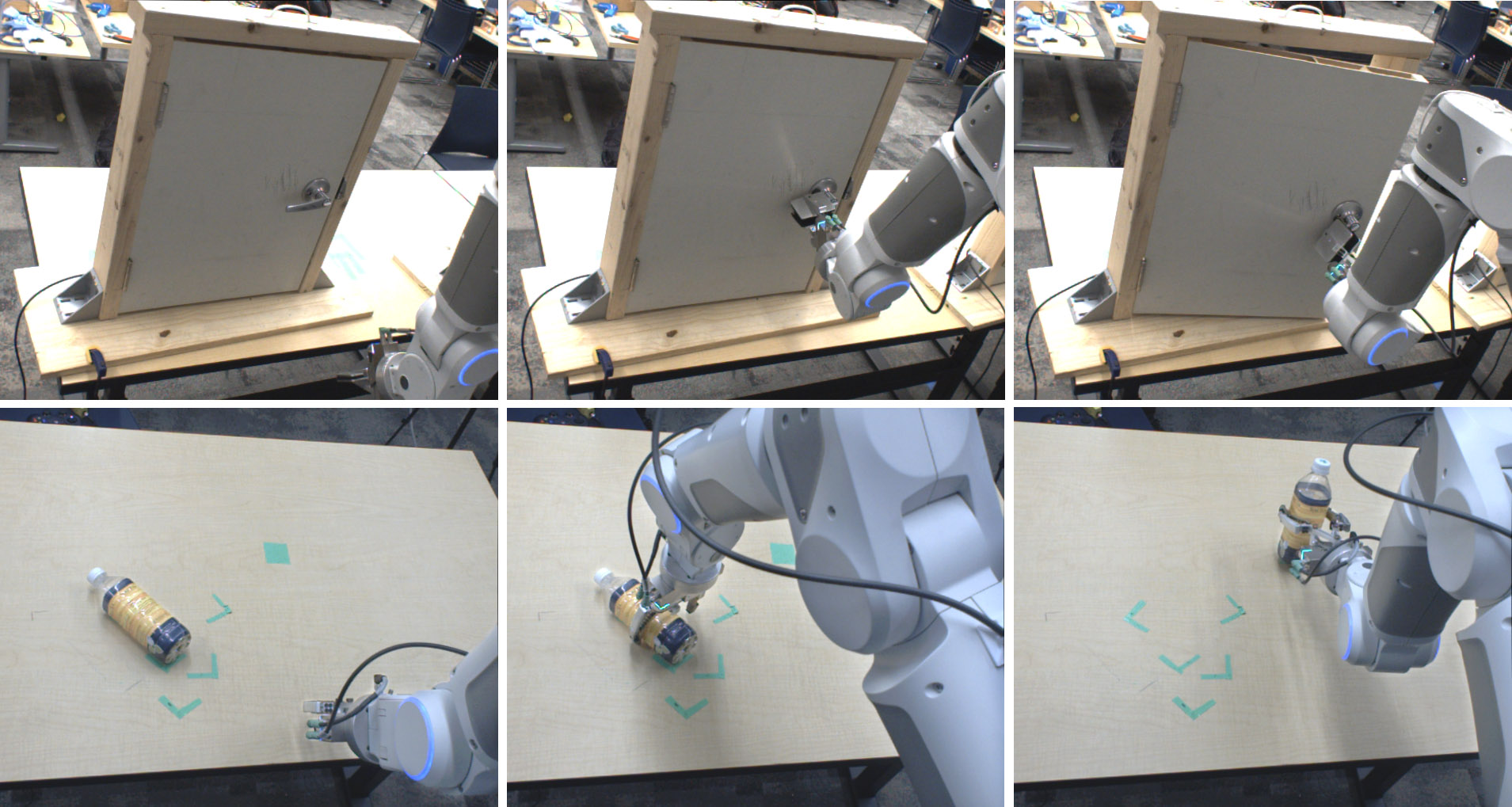}
  \vspace{-19pt}
  \caption{Robot RGB camera images used for controlling the robot. \textit{Top}: door opening task. \textit{Bottom}: pick-and-place task.}
  \label{fig:camera_img}
\end{figure}
In the pick-and-place task adaptation experiment, the goal of the robot is to adapt the demonstrated trajectory to a displaced and rotated object. The bottle is displaced by 5cm and rotated by 30 degrees from its demonstrated position. The local policy consists of 200 time steps. The initial noise is set such that the robot is able to grasp or partially grasp the bottle at its new position at least 10\% of the time.
Figure~\ref{fig:task_adapt} shows performance of PI$^2$ on recovering the pick-and-place behavior for the displaced bottle. 
%REVIEW: point 5
We do not compare this performance to LQR as our cost function could not be modeled with a linear-quadratic approximation.
%END REVIEW: point 5
The robot is able to achieve a 100\% success rate at placing the bottle upright at the target after 8 iterations. The learning is slower than in the door task, and more exploration is required to adapt the demonstration, since the robot has to not only grasp the object but also place it into a stable upright position to succeed. This is difficult, as the bottle might rotate in hand during grasping, or might tip over when the robot releases it at the target. Hence, both stages of the task must be executed correctly in order to achieve success. We noticed that the robot learned to grasp the bottle slightly above its center of mass, which made stable placement easier. Furthermore, this made the bottle rotate in the hand towards a vertical orientation when placing it at the goal position.

\subsubsection{Simulation}
\label{sec:sim_reps}

The goal of this experiment is to compare the use of PI$^2$ with GPS and global policy sampling (called PI-GPS), with an approach based on relative entropy policy search (REPS)~\cite{PetersMA10}, wherein the global policy is directly trained using the samples reweighted by their probabilities $P_{i,t}$ as in~\cite{Hoof0N15}, without first fitting a local policy. We also evaluate a hybrid algorithm which uses PI$^2$ to optimize a local policy, but reuses the per-sample probabilities $P_{i,t}$ as weights for training the global policy (called PI-GPS-W). For these evaluations, we simulate a 2-dimensional point mass system with second order dynamics, where the task involves moving the point mass from the start state to the goal. The state space consists of positions and velocities ($\mathbb{R}^4$), and the action space corresponds to accelerations ($\mathbb{R}^2$). We use a fully connected neural network consisting of two hidden layers of 40 units each with ReLU activations as the global policy. Each algorithm is tuned for optimal performance using a grid search over hyperparameters.

\begin{figure}
  \centering
  \includegraphics[trim=0cm 0 0cm 0, clip=true,width=0.49\columnwidth]{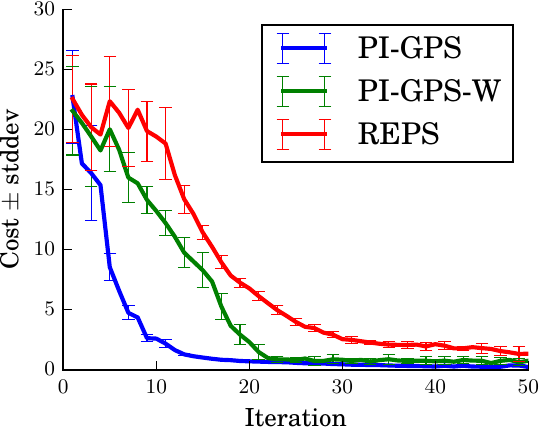}
  \includegraphics[trim=0cm 0 0cm 0, clip=true,width=0.49\columnwidth]{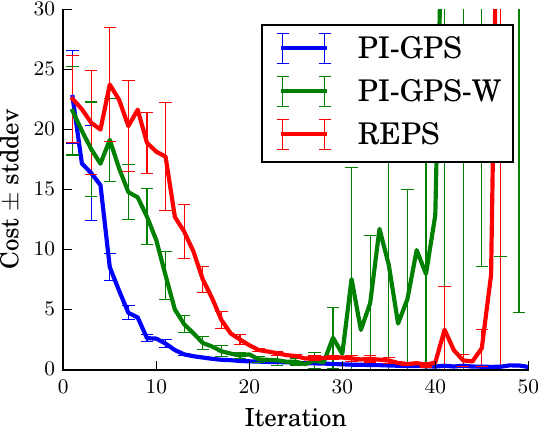}
  \vspace{-18pt}
  \caption{Training curves for PI-GPS, PI-GPS-W, and REPS on a simulated point mass task. \textit{Left}: Runs chosen based on lowest mean cost at iteration 50; \textit{Right}: Runs chosen based on lowest mean cost across all iterations. Each iteration consists of 30 trajectory samples for a single task instance.}
  \label{fig:reps}
  \vspace{0pt}
\end{figure}

The results of this experiment are shown in Figure~\ref{fig:reps}. PI-GPS achieves lower costs at convergence, and with fewer samples. We also observe that PI-GPS-W and REPS have a tendency to go unstable without achieving convergence when training complex neural network policies. We found that this effect disappears when training a linear policy, which may suggest that training a non-linear policy with higher representation power is more stable when the training examples are consistent (like those generated from an optimized local policy in PI-GPS), rather than noisy reweighted samples (like the ones used in REPS).

\subsection{Evaluating generalization}
In this section, we evaluate the generalization performance of guided policy search with PI$^2$ on randomized instances (e.g. random door poses or bottle positions), so as to determine the degree to which global policy sampling improves generalization. We learn visuomotor neural network policies for door opening and pick-and-place tasks that map RGB images from the camera to the torque commands.

\subsubsection{Door opening}
The goal of applying global policy sampling in the door task is to teach the robot to open the door placed at any pose inside the training area.
The variation of the door position is 16cm in $x$ and 8cm in $y$ direction. The orientation variation is $60^{\circ}$ ($\pm30^{\circ}$ from the parallel orientation of the door with respect to the table edge).
Figure~\ref{fig:camera_img} (top) shows examples of the images from the robot's camera during these tasks. Training is initialized from 5 demonstrations, corresponding to 5 door poses. The convolutional layers of the network are trained using 1813 images of the door in different poses, and an additional 15150 images of the robot as described in Section \ref{sec:visuomotor}.

We compare the performance of the global policy trained with the standard local policy sampling method on the demonstrated door poses (as in prior work~\cite{Levine:2016}) to global policy sampling, where new random door poses are sampled for each iteration. We first perform two iterations of standard GPS with local policy sampling to initialize the global policy, since sampling from an untrained neural network would not produce reasonable behavior. In the global policy sampling mode, we then train for 5 iterations with random door poses each time, and compare generalization performance against a version that instead is trained for 5 more standard local policy sampling iterations. In both cases, 7 total iterations are performed, with each iteration consisting of 10 trajectory samples for each of 5 instances (for a total of 50 samples per iteration). PI$^2$ is used to update each instance independently using the corresponding 10 samples. During global policy sampling, we increase the control noise after initialization to add enough exploration to touch the handle of a randomly placed door in at least 10\% of the roll-outs. Figure~\ref{fig:global} (left) shows the average success rates during training.

To test each policy, we evaluate it on 30 random door poses in the training area. When trained using only local policy sampling on the same set of 5 instances at each iteration, the robot is able to successfully open the door in 43.3\% of the test instances. When trained with global policy sampling, with a new set of random instances at each iteration, the final policy is able to open the door in 93.3\% of the test instances.

It is important to note that, during global policy sampling, we had to use relatively low learning rates to train the neural network. In particular, after initialization we had to reduce learning rate by 5 times from $5\times10^{-3}$ to $10^{-3}$. 
Otherwise, we faced the problem of the policy ``forgetting'' the old task instances and performing poorly on them when trained on the new random set of instances each iteration. This could be mitigated in future work by using experience replay and reusing old instances for training.

\subsubsection{Pick-and-place}

%REVIEW: point 6
\begin{figure}
  \centering
  \includegraphics[trim=0pt 0pt 3pt 0pt, clip=true,width=0.99\columnwidth]{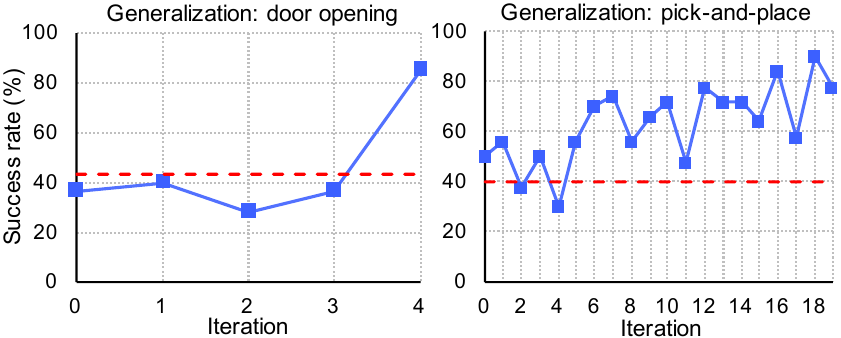}
  \vspace{-20pt}
  \caption{Success rates during training generalized policies with global policy sampling for door opening (\textit{left}) and pick-and-place (\textit{right}). Each iteration consists of 50 trajectory samples: 10 samples of each of the 5 random task instances. %resulting in 50 samples per iteration. 
  \textit{Dashed lines:} success rates after local policy sampling training.}
  \label{fig:global}
  \vspace{-3pt}
\end{figure}
%END REVIEW: point 6

In the pick-and-place global sampling experiments, we teach the robot to pick up a bottle placed at any position inside a predefined rectangular training area of the table and place it vertically at the goal position. The size of the training area is 30x40cm and orientation variation of the bottle is $120^{\circ}$.
Similar to the door opening task, training is initialized from 5 demonstrations, and the convolutional layers are trained using 2708 images with object poses and 14070 task execution images with end-effector poses. Example camera images are shown in Figure~\ref{fig:camera_img} (bottom).

As in the door task, the policy is initialized with 2 local sampling iterations. After that, we run 20 iterations of global policy sampling, each consisting of 10 trajectory samples from each of 5 random instances. During training, we gradually increased the training area by moving away from the demonstrated bottle poses. This continuation method was necessary to ensure that the bottle could be grasped at least 10\% of the time during the trials without excessive amounts of added noise, since the initial policy did not generalize effectively far outside of the demonstration region.

We evaluated the global policy on 30 random bottle poses. After the two initialization iterations, the global policy is able to successfully place the bottle on the goal position with a success rate of 40\%. After finishing the training with global policy sampling, the robot succeeds 86.7\% of the time.  Figure~\ref{fig:global} (right) shows the average success rates over the course of the training. Similar to the task adaptation experiments, the learning of the pick-and-place behavior is slower than on the door opening task, and requires more iterations. Since the size of training region increased over the course of training, the performance does not improve continuously, since the training instances are harder (i.e., more widely distributed) in the later iterations.

We noticed that the final policy had less variation of the gripper orientation during grasping than in the demonstrated instances. The robot exploited the compliance of the gripper to grasp the bottle with only a slight change of the orientation. In addition, the general speed of the motion decreased over the course of learning, such that the robot could place the object more carefully on the goal position. 

During global policy sampling phase, we had to reduce the learning rate of the SGD training   even more than in the door task by setting it to 10$^{-4}$ (compared to $5\times10^{-3}$ for local policy sampling) to avoid forgetting old task instances.

\vspace{2pt}
\section{Conclusions and Future Work}

We presented the path integral guided policy search algorithm, which combines stochastic policy optimization based on PI$^2$ with guided policy search for training high-dimensional, nonlinear neural network policies for vision-based robotic manipulation skills. The main contributions of our work include a KL-constrained PI$^2$ method for local policy optimization, as well as a global policy sampling scheme for guided policy search that allows new task instances to be sampled at each iteration, so as to increase the diversity of the data for training the global policy and thereby improve generalization. We evaluated our method on two challenging manipulation tasks characterized by intermittent and variable contacts and discontinuous cost functions: opening a door and picking and placing a bottle. Our experimental evaluation shows that PI$^2$ outperforms the prior LQR-based local policy optimization method, and that global policy sampling greatly improves the generalization capabilities of the learned policy.

One limitation of our approach is that, since PI$^2$ performs local improvements to the local policies based on stochastic search, the initial structure of the behavior typically needs to be provided through human demonstrations. However, with global policy sampling, we can supply demonstrations for only a few instances of the task, and still benefit from training on a wide range of random instances. A promising avenue for future work is to explore ways to combine the strength of the LQR-based optimizer, which can make large changes to the local policies based on estimated gradients, with the capability of PI$^2$ to finely optimize the policy in the presence of challenging cost and dynamics discontinuities.

\section*{Acknowledgements}
We would like to thank Peter Pastor and Kurt Konolige for additional engineering, robot maintenance, and technical discussions, and Ryan Walker and Gary Vosters for designing custom hardware for this project.

\bibliographystyle{unsrtnat}
\renewcommand{\bibfont}{\footnotesize}  
\bibliography{PIGPS}
\end{document}